# Event-Conditioned Diagnostics of Kinematic, Contact, and Object-Permanence Structure in Passive Object-State World Models

Yang Liu, Yuming Chen
College of Intelligent Robotics and Advanced Manufacturing, Fudan University
Shanghai, China
{ly, yumingchen}@fudan.edu.cn

**Abstract**

World models can predict future physical states, but prediction accuracy alone does not explain how physical information is organized and used inside their latent dynamics. We introduce a controlled diagnostic protocol for studying event-conditioned latent physical structure in passive object-state world models. The protocol separates three questions: whether event-regime information is readable, whether event context changes the relative emphasis of kinematic-, contact-, and object-permanence-related readouts, and whether suppressing readout-aligned directions affects prediction. Using a balanced controlled-generator dataset with free-motion, collision, and occlusion events, we evaluate GRU, Transformer-lite, and RSSM-lite transition models under a fixed-horizon forecasting setup. Across three independent seeds, all three architectures learn useful predictive dynamics and support reliable event-regime readout. Event context systematically shifts the relative readout pattern: free motion is kinematic-dominant, collision increases contact-related structure, and occlusion increases object-permanence-related structure. Time-aligned analyses show corresponding phase-related shifts. Fixed-horizon projection Causal Field Effect (CFE), used here as a functional-sensitivity test, gives the clearest evidence for contact-aligned structure in collision-contact windows. Object-permanence-aligned structure is also sensitive during hard occlusion, but control specificity remains mixed. These results support event-conditioned latent structure and functional sensitivity without implying explicit physical modules or isolated causal circuits.

## 1. Introduction

World models learn latent dynamics that support prediction, planning, and control (Ha & Schmidhuber, 2018; Hafner et al., 2019; Hafner et al., 2023; Micheli et al., 2022). Yet strong prediction does not tell us how physical information is organized inside the hidden state. Variables such as motion direction or contact may be distributed rather than represented as explicit coordinates (Joseph et al., 2026). We therefore ask not only whether a model predicts correctly, but how physical information is organized and used in its latent dynamics.

We study event-conditioned organization: the same model may emphasize different physical information as the event changes. We treat kinematic, contact, and object-permanence labels as overlapping diagnostic categories, not separate internal modules. Collision still contains kinematic structure before contact, and occlusion still contains visible motion before hiding. The question is whether the relative emphasis changes when contact or hidden-state ma[1]intenance becomes important.

---

Source code, data, and reproduction materials are available at

https://github.com/lysea8282/event-conditioned-world-model-diagnostics

Benchmarks measure behavior, object-centric models improve prediction, and probing or steering studies show that physical information can be readable or manipulable. Nevertheless, what is less clear is how these properties change across event contexts in one controlled setting. Our protocol therefore tests event-regime readability, event-conditioned readout reweighting, phase/time consistency, and fixed-horizon functional sensitivity under projection suppression.

## 2. Related Work

### 2.1 Physical reasoning benchmarks

Physical reasoning benchmarks make physical success and failure measurable. Physion evaluates visual physical prediction across collisions, stability, rolling, sliding, and projectile motion (Bear et al., 2021). CLEVRER focuses on collision events and asks descriptive, explanatory, predictive, and counterfactual questions about videos (Yi et al., 2019). IntPhys and IntPhys2 test violation-of-expectation judgments for intuitive-physics principles such as permanence, continuity, and solidity (Bordes et al., 2025). CATER stresses compositional actions, temporal reasoning, and object tracking under controlled scene biases (Girdhar & Ramanan, 2019). PHYRE evaluates sample-efficient physical problem solving in 2D mechanics puzzles (Bakhtin et al., 2019). These benchmarks define observable behavioral criteria for physical reasoning. But they mainly evaluate task behavior, plausibility judgments, causal question answering, or planning outcomes. They do not by themselves explain how physical information is organized inside a model.

### 2.2 Object-centric dynamics and world models

Object-centric dynamics models decompose scenes into objects, slots, particles, or relational state variables, and then learn how these entities evolve over time (Shamsian et al., 2020; Song et al., 2025; Villar-Corrales et al., 2023; Wu et al., 2022). They can improve long-horizon prediction, future reasoning, visual question answering, planning, and object tracking. Some models directly target object permanence and occlusion. They learn to track hidden objects, anticipate reappearance, or maintain object-specific state through occlusion (Shamsian et al., 2020; Traub et al., 2024). Other models introduce explicit kinematic structure or separate object dynamics from object interactions (Song et al., 2025; Villar-Corrales et al., 2023). This line of work shows that object structure, kinematic structure, and interaction structure can improve prediction and generalization. Its main goal, however, is usually better rollout, better prediction, better planning, or better generalization. It is less often focused on how latent physical structure changes its organization and functional role when the physical event changes.

### 2.3 Probing, steering, and representation diagnostics

Internal representation analysis comes closer to our diagnostic focus. Recent interpretability studies show that physical variables can be decoded from hidden states of video or world models. Speed, acceleration, motion direction, object position, and game-state variables can become accessible in learned representations (Joseph et al., 2026; Zhang, 2026). These representations are not always simple factorized variables. For example, direction can be encoded through a distributed population geometry rather than a single explicit coordinate (Joseph et al., 2026). Other work goes beyond probing. Probe-derived Concept Activation Vectors can steer physical plausibility judgments in video world models (Alam, 2026). Interventions along probe-derived directions can also change the predictions of learned environment simulators (Zhang, 2026). Causal and contextual world-model studies further show that learned latent structure can support counterfactual dynamics, intervention-aware prediction, or context-dependent rollout behavior (Ding et al., 2026; Lei et al., 2022; Röder et al., 2026). These results show that physical information can be readable, geometrically organized, and sometimes functionally used.

## 3. Problem formulation

### 3.1 Passive Object-State World Models

In this work, we study passive object-state world models, where a model observes a sequence of object states and predicts how those states evolve over time. Each state contains object-level variables such as position, velocity, size, visibility, or other available annotations. We write the object-state sequence as $s_{1:T}$, where $s_t$ summarizes the scene at time t. A transition model is the part of the world model that maps observed states to future states. It encodes the observed prefix into hidden states $h_t$, and predicts a future trajectory $\hat{s}_{t+1:t+H}$, where H is the prediction horizon. In our experiments, this prefix-horizon interface is fixed: the models observe the first eight frames and predict the next eight frames. Although this fixed-horizon setup does not test arbitrary sliding-context generalization, it provides a controlled forecasting interface in which all architectures receive the same observed prefix and are evaluated over the same future target horizon. This design makes the subsequent diagnostic comparisons more interpretable by minimizing potential confounds from differences in context length, prediction horizon, and event-aligned resampling.

The word passive means that the models considered here do not receive agent actions as input. They are not trained to choose actions, plan interventions, or control a robot. They only learn the dynamics of observed physical scenes, which makes the setting different from action-conditioned reinforcement-learning world models. Therefore, our diagnostic target is narrowed.

We use object-state inputs rather than raw pixels. This choice removes the confound of object discovery, segmentation, and visual feature extraction. It allows us to focus on the transition dynamics themselves. The goal of this article is not to test whether a model can perceive objects from video, but to test whether, given object-level state information, the transition model organizes latent physical structure in a way that depends on the physical event context.

Although our setting is passive, it is still relevant to more general action-conditioned and control-oriented world models. Such models also rely on latent states to predict future dynamics, evaluate possible futures, or support downstream planning and control. In those settings, action selection depends on how the model represents the physical consequences of states and actions. Though a passive transition model does not choose actions, it provides a controlled setting for a prerequisite question: whether the latent state contains event-sensitive physical structure at all.

### 3.2 Physical Event Contexts

We use physical event context to describe the type of physical situation shown by an observed trajectory segment. An event context is not necessarily a label given to the model as input. It is a diagnostic grouping used. It helps us ask what kind of physical structure the model must represent in order to predict the sequence. In this paper, we focus on three event families: free motion, collision, and occlusion.

A free-motion context is a segment in which objects move without contact or disappearance. Prediction in this regime mainly depends on kinematic structure, such as position, velocity, and smooth motion. A collision context is a segment in which objects come into contact and their motion changes after interaction. Prediction in this regime requires sensitivity to contact, interaction, and post-contact state changes. An occlusion context is a segment in which an object becomes partially or fully hidden and later reappears. Prediction in this regime requires the model to maintain object-specific state across temporary invisibility.

These event families do not cover all of physical reasoning. They are chosen because they isolate three common demands on a physical prediction model. Free motion stresses kinematic continuation. Collision stresses interaction and contact-sensitive dynamics. Occlusion stresses object permanence. Together, they provide a controlled setting for asking whether different physical contexts reweight different latent physical structures.

The event families do not map one-to-one to the diagnostic categories. Collision includes kinematic approach before contact, and occlusion includes visible motion before hiding. We test whether the relative importance of different latent structures changes with event regime.

### 3.3 Latent Physical Structure

We use latent physical structure to mean a pattern in hidden states associated with a physical variable or event demand. The structure may be distributed across many hidden dimensions and may appear as a direction, subspace, or activation pattern. We do not assume that it is an explicit module or a unique physical variable.

For analysis, we use three operational labels: kinematic, contact, and object permanence. Kinematic structure is associated with smooth motion and trajectory continuation; contact structure with interaction and post-contact change; and object-permanence structure with maintaining object state across temporary invisibility.

These labels define diagnostic readouts, not supervised modules. The transition models are not trained with these diagnostic labels. We test whether the corresponding readouts change with event context and whether suppressing readout-aligned directions changes event-relevant prediction. Readability alone is not enough. A useful diagnostic structure should also show context sensitivity and functional relevance.

### 3.4 State-Conditioned Structure Diagnostics

We define event-conditioned structure diagnostics as tests relating three things: event context, hidden-state structure, and prediction behavior. The goal is not simply to decode one variable, but to ask whether the role of latent structure changes with the event.

Standard probes test what information is readable. Our protocol asks the stronger question of whether the relevant structure becomes more prominent when it matters and whether perturbing it changes prediction. We analyze three levels. Event-level diagnostics test whether free motion, collision, and occlusion are separable in hidden representations. Phase-level diagnostics use dataset annotations to describe how physical demands change within an event. Timewise diagnostics use per-time-step hidden states when the architecture exposes comparable trajectories. Phase labels are used only for analysis, not as model inputs.

These levels have different evidential strength. Event-level results provide the main cross-architecture evidence. Phase-level proxies explain the changing physical demands of collision and occlusion. Timewise hidden-state results provide stronger temporal evidence where comparable hidden trajectories are available.

Our goal is not to prove that the model contains explicit physical modules or human-defined physical laws. The goal is to obtain controlled evidence about how physical structure is organized, reweighted, and used inside passive transition dynamics.

## 4. Diagnostic Protocol

### 4.1 Overview

Our protocol checks three main points: separability, selectivity, and functional relevance. Separability asks whether hidden states distinguish event regimes. Selectivity asks whether event context changes the relative emphasis of the diagnostic readouts. Functional relevance asks whether suppressing a readout-aligned direction changes event-relevant prediction within the fixed future horizon.

Phase-level and timewise analyses show how these patterns change over an event. Phase-level proxies come from annotated object states and event labels; timewise analyses come from model hidden states. We keep these two sources of evidence separate.

The protocol is diagnostic, not a full mechanistic explanation. Probe success shows readability, selectivity shows structured association, and CFE adds intervention evidence. None of these alone identifies a complete mechanism.

### 4.2 Event Probe

The first part of the protocol asks whether the model's hidden states separate different physical events. Given a hidden state $h_t$, we train a lightweight probe to classify whether the sequence is free motion, collision, or occlusion. The event label is used only for diagnosis. It is not given to the transition model as an input.

The event probe asks only whether event-regime information is readable from the hidden representation. Probe success by itself does not show that the corresponding structure is functionally used.

### 4.3 Field Selectivity

The second part tests whether event context changes the relative emphasis of the operational readouts. The mapping is not one-to-one: free motion should be mainly kinematic, collision should retain kinematic structure while increasing contact-related structure, and occlusion should retain motion-related structure while increasing object-permanence-related structure during hidden or reappearance phases.

We retain the term field selectivity as an operational name for this readout-based diagnostic; here, "field" refers only to the diagnostic label, not to an assumed internal physical field. It asks whether the relevant readout is stronger in the event where that information should matter. This goes beyond ordinary variable probing because it tests context-dependent emphasis rather than readability alone.

### 4.4 Causal Field Effect (CFE)

The third part asks whether a readout-aligned pattern matters for prediction. CFE measures the change in event-window prediction loss after suppressing that direction. A positive effect shows functional sensitivity in the configured fixed-horizon setting; it does not identify a complete causal mechanism. The "field" in this name refers only to the operational readout label, not to an assumed internal physical field.

For example, if contact-aligned structure is functionally relevant, suppressing a contact-related pattern should affect collision predictions more than free-motion predictions. If object-permanence-aligned structure is functionally relevant, suppressing an object-permanence-related pattern should affect hard-occlusion predictions. We compare target suppression with matched controls, including random directions, random subspaces, mismatched diagnostic labels, and field-label shuffles. These controls help distinguish target-related sensitivity from general perturbation sensitivity.

### 4.5 Controls and Input Variants

The protocol includes controls for shortcut and leakage. If the model receives direct event labels, contact labels, future states, or hidden true states as input, then a probe or readout diagnostic may succeed for trivial reasons. We therefore test whether the diagnostic signals persist after removing obvious shortcut information from the input. First, we use a motion-only setting in which the model receives only object-level position and velocity, without explicit event, contact, or occlusion labels. Second, for occlusion sequences, we use an observation-like setting in which the model receives only visible state information and visibility indicators, without access to the true hidden state of the occluded object. If the diagnostic signals remain under these restricted settings, they are less likely to be explained by direct label leakage or trivial input cues.

## 5. Dataset and Models

### 5.1 Main Controlled Public-Generator Dataset

Our main experiments use a Kubric-style public-generator pipeline to construct an object-state dataset(Greff et al., 2022). The dataset is designed for diagnostic analysis, not as a general-purpose video benchmark. In our dataset, physical event types are balanced, object-level states are available, and diagnostic labels can be separated from model inputs.

The main dataset contains three event families: free motion, collision, and occlusion. Each family contains 300 sequences, giving 900 sequences in total. Free-motion sequences test kinematic continuation. Collision sequences test contact-sensitive interaction. Occlusion sequences test object permanence when an object becomes temporarily hidden. This balanced design lets us compare event-conditioned latent structure across physical regimes without making the analysis dominated by a single event type.

The dataset provides object-level information and diagnostic annotations. These include object states, masks, centroid heatmaps, pairwise relations, event family labels, event phase labels, and metadata. Collision sequences are audited to ensure that they contain true contact events, rather than only near-contact cases. Occlusion sequences are divided into easy and hard cases.

The dataset also contains held-out stress-test splits. For collision, the held-out split emphasizes more difficult collision regimes, such as larger collision angles and higher approach-speed conditions. For occlusion, the held-out split emphasizes longer hard-occlusion windows. We refer to these as out-of-distribution, or OOD, tests, which means testing on harder parameter regimes within the same controlled generator setting.

**5.2 Event Families and Annotations**

The dataset contains three event families: free motion, collision, and occlusion. These event families are not meant to cover all possible physical situations. They are chosen because they place different demands on a passive object-state prediction model.

In free-motion sequences, objects move without contact and without becoming hidden. In collision sequences, two or more objects come into contact and their motion changes after interaction. In occlusion sequences, an object becomes partially or fully hidden and later reappears. The dataset also records event phases such as pre-contact, contact, post-contact, pre-occlusion, hidden, and reappearance. These phase labels are used for analysis and auditing. They are not treated as ordinary model inputs.

This annotation design allows us to compare free motion, collision, and occlusion under a shared object-state format. It also allows us to distinguish event-level diagnostics from phase-level proxy analysis.

**5.3 Transition Models**

We evaluate the diagnostic protocol on three passive object-state transition models: a GRU-based recurrent model, a Transformer-lite model, and an RSSM-lite model. These models serve as architecture-family controls. The purpose is to test whether the diagnostic results depend on one particular transition architecture, or whether similar event-conditioned patterns can be observed across different ways of modeling temporal dynamics.

The GRU-based model represents a recurrent transition architecture. The Transformer-lite model represents an attention-based transition architecture. The RSSM-lite model represents a latent state-space style transition architecture. We do not claim that the RSSM-lite variant is equivalent to a full Dreamer-style agent. We use it only to test whether the diagnostic protocol remains meaningful in a state-space transition setting.

For all three models, the diagnostic target is the hidden representation produced during passive prediction. We train the models to predict future object states. We then analyze their hidden states using event probes, field selectivity, causal field effect, and timewise readouts when appropriate.

### 5.4 Input Variants and Leakage Control

We evaluate several input variants. The first variant uses object-state information for passive prediction. The second variant is a kinematic-only input, which uses only basic motion variables such as object position and velocity. The third variant is an observation-like input, which is designed for occlusion. It uses visible state information, the last observed position, and a visibility flag, without giving the model the true hidden position of an occluded object.

A central risk in event-conditioned diagnostics is shortcut learning. If the model directly receives event labels, contact labels, future states, or hidden true states as input, then the diagnostic may succeed for a trivial reason. To avoid this problem, event family labels, event phase labels, contact labels, future states, and hidden true object states are absent from context inputs. They are used only to define diagnostic groups, check dataset quality, and compute evaluation metrics.

### 5.5 Training and Diagnostic Readouts

All transition models are trained for passive future-state prediction. Given an observed prefix of object states, the model predicts future object states. The training objective is prediction loss on future object states. The models are trained separately for each architecture, input variant, and random seed. The main evaluation uses the in-distribution test split. Held-out collision and occlusion splits are used later for stress tests. In the reported experiments, the primary training and evaluation interface is fixed-horizon: an observed prefix is used to predict a configured future horizon.

After training, we analyze the hidden states of the trained transition models. The diagnostic readouts are fitted after the transition model has learned the prediction task. They are used only for analysis, not used to train the transition model, and their labels are not given to the model as input.

The first readout is an event probe. Both standard variable probes and our event probe take model hidden states as input. The difference lies in the prediction target. A standard variable probe usually predicts a specific physical variable, such as position, velocity, contact, or visibility; while our event probe predicts the physical event regime of the sequence: free motion, collision, or occlusion.

Field selectivity and CFE ask stronger questions than the event probe. Field selectivity tests whether readout emphasis changes with event context, and CFE tests whether suppressing a readout-aligned direction changes event-relevant prediction. Exact architecture sizes, optimizer settings, training epochs, normalization details, and intervention implementation are provided in Appendix A.

## 6. Main results

### 6.1 Models Learn Predictive Dynamics and Encode Event-Regime Information

Before analyzing latent physical structure, we first check whether the models learned the basic predictive dynamics. Table 1 reports this check under the kinematic-only input setting. Across all three architectures and three independent seeds, every model predicts future object states better than the constant-velocity baseline on the in-distribution test split.

We then apply the event probe defined in Section 4.2. Prior work has already shown that physical variables can often be decoded from learned representations. Here, the probe mainly checks whether, in our trained passive object-state models, the hidden states contain stable information about the current event regime.

As shown in Table 1, event-regime information is reliably readable from the hidden states across architectures and seeds. We do not treat this as evidence of functional structure by itself. The stronger evidence comes from context-dependent readout changes and intervention-based sensitivity.

Table 1 Prediction sanity and event-regime readout under kinematic-only input. Seed denotes the random seed used for model training and evaluation. MSE denotes normalized future-position MSE over predicted object x/y coordinates in the fixed-horizon target window. CV_MSE is the same metric computed for the constant-velocity baseline. Event probe F1 is the macro-F1 score of the lightweight probe that predicts free motion, collision, or occlusion from frozen model hidden states.

| Model | Seed | MSE | CV_MSE | MSE/CV | Probe F1 | Verdict |
|---|---|---|---|---|---|---|
| GRU | 3407 | 2.46E-04 | 3.35E-04 | 0.73 | 1.00 | Pass |
| Transformer | 3407 | 9.98E-05 | 3.35E-04 | 0.30 | 1.00 | Pass |
| RSSM | 3407 | 1.49E-04 | 3.35E-04 | 0.45 | 1.00 | Pass |
| GRU | 3408 | 2.23E-04 | 3.35E-04 | 0.66 | 1.00 | Pass |
| Transformer | 3408 | 1.21E-04 | 3.35E-04 | 0.36 | 1.00 | Pass |
| RSSM | 3408 | 1.40E-04 | 3.35E-04 | 0.42 | 1.00 | Pass |
| GRU | 3409 | 2.26E-04 | 3.35E-04 | 0.67 | 1.00 | Pass |
| Transformer | 3409 | 9.99E-05 | 3.35E-04 | 0.30 | 1.00 | Pass |
| RSSM | 3409 | 1.45E-04 | 3.35E-04 | 0.43 | 1.00 | Pass |

### 6.2 Event Contexts Reweight Non-Exclusive Physical Structure

The event probe shows that hidden states contain event information. We next ask how that information is organized. We do not assume that one event corresponds to one structure; instead, we test how the relative emphasis of the diagnostic readouts changes across events. Collision still contains ordinary motion before contact, and occlusion still contains visible motion before hiding. We therefore interpret the results as relative reweighting, not exclusive event-to-structure assignment.

Figure 1 summarizes this reweighting at event, phase, and timewise levels. Event-level and timewise model figures use hidden-state readouts. The phase-level figures use annotated data proxies to describe the physical demands of the trajectory itself.

Figure 1A, D, and E report softmax probabilities from a three-class linear readout applied to z-scored hidden states. Figure 1B and C report data-level proxy z-scores, not model activations. Figure 1F reports the fraction of GRU and Transformer-lite cases with shifts in the expected direction. These quantities should be compared within each Figure 1 subfigure, not across subfigures.

At the event level, RSSM readouts show the expected ordering. Free motion is strongly kinematic-dominant; collision keeps a substantial kinematic component while increasing contact-related structure; and occlusion keeps a smaller motion-related component while strongly increasing object-permanence-related structure. The main point is the change in relative emphasis across event contexts.

The phase-level proxy figures explain why collision and occlusion are mixed events. In collision, the proxy pattern shifts from kinematic-dominant pre-contact motion to contact-sensitive interaction at contact, then partly returns toward a kinematic regime. In occlusion, the pattern shifts from visible motion before hiding to object-permanence-dominant structure during the hidden interval, with partial recovery of visible-state structure at reappearance.

RSSM future-rollout latents show a similar temporal pattern. Contact-related readout scores rise around contact while kinematic scores fall; object-permanence scores rise around hidden-start while kinematic scores fall. The strict phase-aware RSSM readout reaches a mean test macro-F1 of 0.896 ± 0.004 across three seeds, supporting time-resolved event-related structure in the RSSM representation.

GRU and Transformer-lite do not expose directly comparable future rollout latent trajectories, so we use a lighter sliding-prefix analysis by repeatedly encoding prefixes ending at successive event phases. Across both input variants and three seeds, the expected contact and object-permanence shifts remain positive in every evaluated case, giving a joint directional consistency score of 1.000.

Together, Figure 1 supports a simple hierarchy: event information is readable, event context changes relative readout emphasis, and the expected temporal shifts also appear in hidden-state analyses. The evidence supports event-conditioned latent structure, not explicit or isolated physical modules.

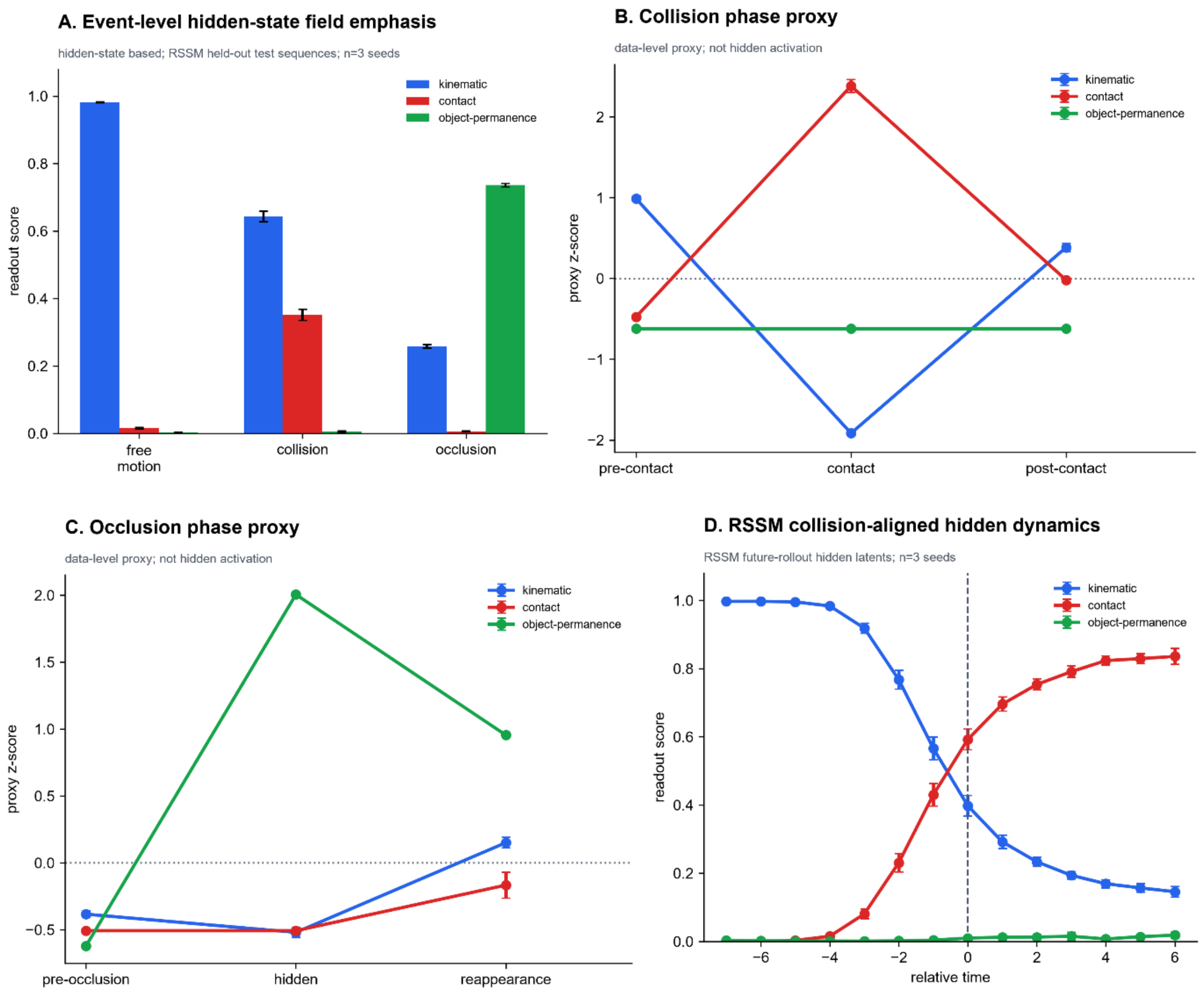

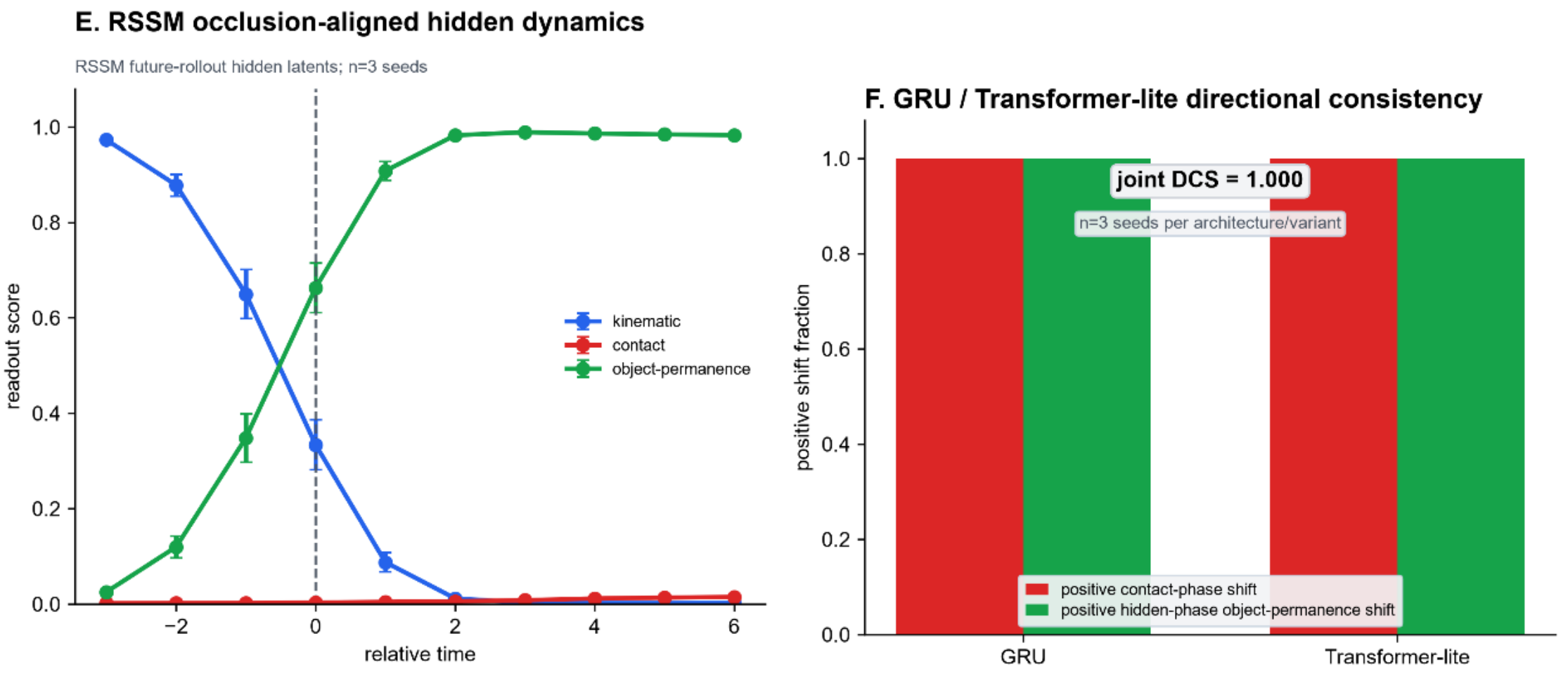

Figure 1 Event contexts reweight non-exclusive latent physical structure at event, phase, and timewise levels. A, D, and E report hidden-state readout probabilities; B and C report data-level proxy z-scores; F reports directional consistency. Values should be compared within each Figure 1 subfigure rather than across subfigures. (A) RSSM event-level readouts show kinematic dominance in free motion, increased contact-related structure in collision, and strong object-permanence-related structure in occlusion. (B-C) Data-level phase proxies show the expected shift toward contact at collision and toward object permanence during occlusion. (D-E) Time-aligned RSSM readouts show analogous shifts around contact onset and hidden-start. (F) GRU and Transformer-lite sliding-prefix analyses retain the expected shift direction across all evaluated cases, with joint DCS = 1.000. The figure supports event-conditioned reweighting of diagnostic structure, not one-to-one event modules.

### 6.3 Fixed-horizon Windowed Projection CFE

Sections 6.1 and 6.2 establish readability and event-conditioned organization. We next test functional sensitivity: does suppressing a readout-aligned component worsen prediction in the event window where it is expected to matter? We answer this question with CFE. CFE is the increase in normalized x/y position prediction loss after projection suppression. It is evaluated only on event-relevant frames inside the fixed prediction horizon. Appendix C defines the windows, readout-aligned directions, intervention, controls, and aggregation.

Figure 2A shows the fixed-horizon event-window coverage. The analysis uses trained checkpoints and does not retrain the world models. For GRU and Transformer-lite, perturbed hidden states are decoded through the prediction head. For RSSM-lite, CFE is a head-decoding test from the perturbed latent state, not a full transition re-rollout. We therefore interpret CFE as fixed-horizon representational sensitivity. No reappearance entries are available in Figure 2A because reappearance falls outside the fixed prediction horizon.

Collision-contact gives the clearest result. At full suppression, contact-aligned CFE is 0.00317 ± 0.00056 (mean ± SEM), with positive effects in 9/9 architecture-seed cases and positive dose trends in 9/9 cases (as shown in Figure 2B and D). Mean target-control margins are also positive against matched random direction (0.00182), random subspace (0.00140), and field-label shuffle (0.00194), as shown in Figure 2C.

Hard occlusion also shows positive sensitivity, but specificity remains mixed. Object-permanence-aligned CFE is 0.00711 ± 0.00084, with positive effects and positive dose trends in 9/9 cases, (as shown in Figure 2B and D). The mean margin is positive against random subspace (0.00229) and field-label shuffle (0.00589), but negative against the matched random direction (-0.00134), as shown in Figure 2C. We therefore have a conservative interpretation: the object-permanence-aligned direction is functionally sensitive during hidden windows, but it is not uniquely specific across all controls.

CFE therefore adds functional evidence beyond probing, especially for collision-contact. The controls also show why a positive target effect is not enough on its own. Still, these interventions do not mean the models have explicit modules, isolated causal circuits, or a complete mechanistic decomposition related to the events.

Overall, the evidence supports fixed-horizon functional relevance of event-conditioned latent structure. Contact-aligned structure has the strongest and most consistent specificity evidence; object-permanence-aligned structure shows positive sensitivity with mixed control specificity.

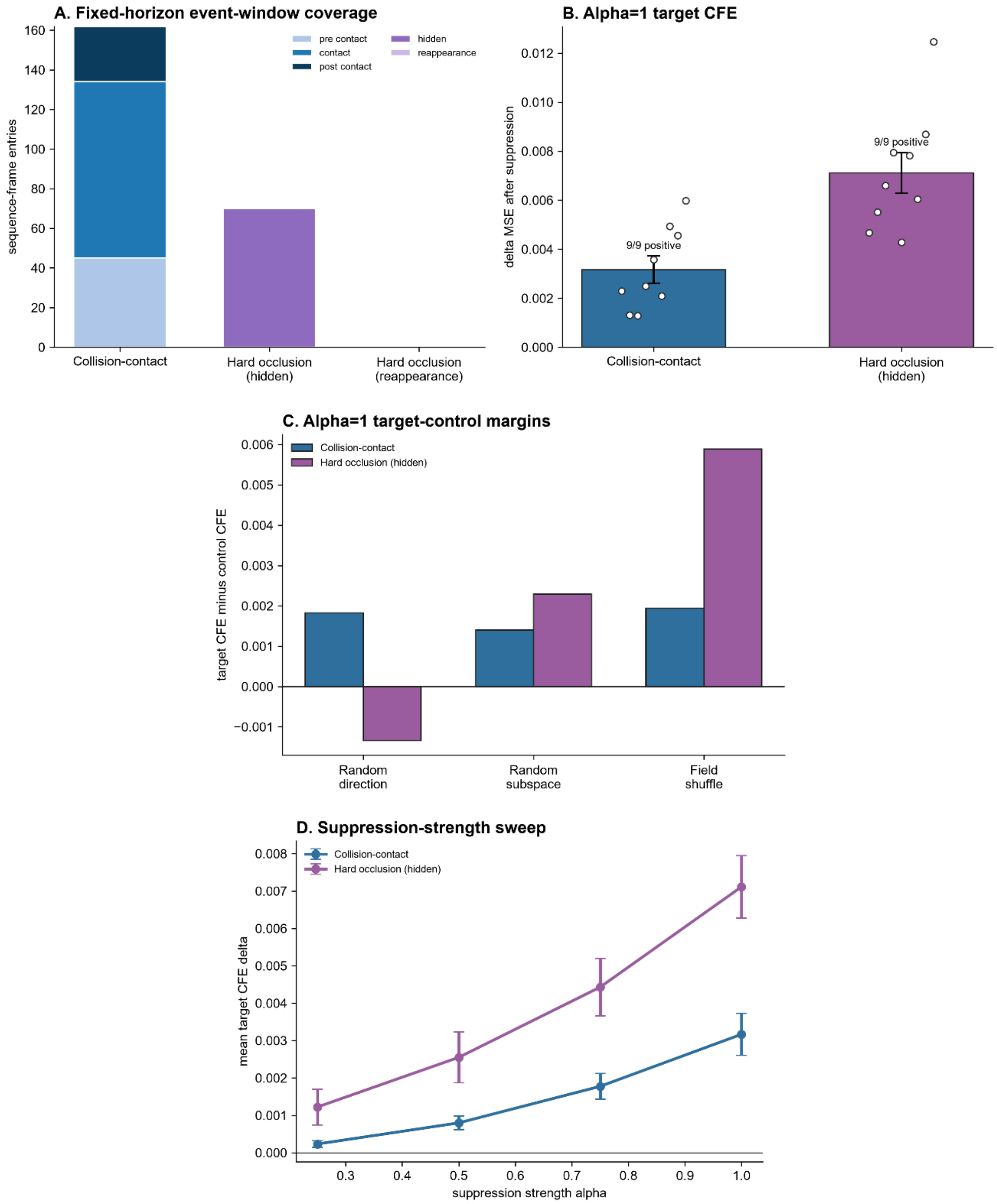

Figure 2 Windowed projection CFE supports fixed-horizon functional sensitivity. (A) Event-window coverage for collision-contact and hard-occlusion hidden windows. (B) Mean target CFE at full suppression ($\alpha = 1$) over nine architecture-seed cases per event; both targets are positive in 9/9 cases. (C) Target-control margins at $\alpha = 1$. Collision-contact is positive against all reported controls. Hard occlusion is positive against random-subspace and field-shuffle

controls but negative against the matched-random-direction control. (D) Suppression-strength sweep; all nine cases in each event show a positive target dose trend. The figure supports fixed-horizon functional sensitivity, not explicit physical modules or full rollout-level causal mechanisms.

## 7. Discussion and Limitations

### 7.1 Contribution of our diagnostic protocol

The main contribution of this article is a diagnostic protocol for asking how latent physical structure is organized and used under different event contexts. Prediction metrics show whether a model predicts well, and probes show what information is readable. Our protocol adds context-dependent organization and intervention-based functional sensitivity. The protocol separates three evidence levels: readability, event-conditioned organization, and functional sensitivity. This matters because physical information can be present in a hidden state without being used in the same way across events. Because the protocol operates on hidden representations rather than architecture-specific modules, it can compare recurrent, attention-based, and state-space transition models under matched event conditions.

### 7.2 Limitations & future extensions

The main limitation is the controlled fixed-horizon setting. In the reported experiments, models observe the first eight frames and predict the next eight frames. This design makes cross-architecture comparisons more interpretable, because all models receive the same observed prefix and are evaluated over the same future horizon. However, it does not establish arbitrary sliding-context generalization. A model that exhibits event-conditioned structure under this fixed forecasting interface may behave differently under event-aligned sliding windows, longer horizons, or online rollout.

A second limitation is the use of object-state inputs rather than raw video. This removes object discovery and visual perception from the experiment, letting us focus on transition dynamics. The results therefore do not show that raw-video world models will learn the same latent structure.

A third limitation concerns CFE. Projection suppression targets a readout-aligned direction, not an explicit module. For RSSM-lite, the reported CFE is head-decoding sensitivity rather than a full transition re-rollout. Hard-occlusion sensitivity is also limited to hidden-window prediction, and specificity remains mixed because the target does not exceed every matched control. These limits keep the functional claim narrow.

The event taxonomy is also limited. Free motion, collision, and occlusion cover kinematic continuation, contact-sensitive interaction, and object permanence, but not many other physical processes. We therefore treat these three categories as a minimal diagnostic set, not a complete taxonomy of physical understanding.

Several extensions are natural: longer or sliding-context forecasting, full rollout interventions, richer physical events, and action-conditioned settings. These extensions should preserve the same separation between readability, organization, and functional sensitivity. A broader extension is raw-video and action-conditioned world models. The present object-state setting is a controlled first step that isolates transition dynamics before adding perception and action.

## 8. Conclusion

We introduced a controlled diagnostic protocol for event-conditioned latent physical structure in passive object-state world models. The protocol separates what is readable from how structure is reorganized by event context and whether readout-aligned perturbations affect prediction. Across three architectures and three independent seeds, the models learn useful fixed-horizon dynamics and reliable event-regime readouts. Event context changes the relative emphasis of kinematic-, contact-, and object-permanence-related structure. Projection CFE provides the strongest functional evidence for contact-aligned structure in collision-contact windows, while object-permanence-aligned structure shows positive hard-occlusion sensitivity with mixed control specificity. The results support a reusable diagnostic approach to latent physical structure without implying explicit modules or isolated causal circuits.

**Declaration of generative AI and AI-assisted technologies in the manuscript preparation process**

During the preparation of this work, the authors used ChatGPT and OpenAI Codex to assist with literature organization, experimental and code development, analysis organization, and manuscript editing. The authors reviewed and edited all AI-assisted output and take full responsibility for the content of the article.

## Appendix A. Experimental and Diagnostic Details

This appendix documents the fixed-horizon scale/OOD evaluation used as the methodological baseline for the manuscript. The goal is to make the controlled public-generator experiment, model training, hidden-state extraction, event probing and Causal Field Effect (CFE) audit history reproducible. The original baseline CFE sweep documented here is retained as baseline and audit information; the primary fixed-horizon windowed projection CFE results reported in Section 6.3 and Figure 2 are defined separately in Appendix C.

### A.1 Dataset generation

The fixed-horizon scale/OOD evaluation generated the controlled robustness dataset physem_publicgen_balanced_v2_scale_ood. The dataset contains three event families: free motion, collision, and occlusion. Each event family contains 300 sequences, for a total of 900 sequences. Each sequence has 30 frames, two object slots, 64 × 64 rendered frames, object masks, centroid heatmaps, pairwise relation tensors, event-family labels, event-phase labels, sequence identifiers, and metadata. Table A.1 summarizes the fixed dataset configuration.

**Table A.1.** Controlled-generator dataset configuration for the fixed-horizon scale/OOD evaluation.

| Item | Value |
|---|---|
| experiment_id | fixed_horizon_scale_ood_evaluation_v0 |
| dataset | physem_publicgen_balanced_v2_scale_ood |
| n_sequences | 900 |
| n_per_event | 300 |
| events | free_motion, collision, occlusion |
| sequence_length_T | 30 |
| resolution | 64 × 64 |
| objects_K | 2 |
| context_length | 8 |
| future_horizon | 8 |
| dataset_seed | 7707 |

The object-state fields are: x, y, $v_x$, $v_y$, visible, occluded, object_id, radius, mass, shape_id. The pairwise-relation fields are: distance, near_contact, contact, relative_$v_x$, relative_$v_y$, collision_phase. The stored tensor set is: frames, object_states, object_masks, centroid_heatmaps, pairwise_relations, event_family, event_phase, metadata_json.

Collision sequences were generated with true-contact dynamics rather than near-contact-only templates. The dataset audit reports a collision true-contact rate of 1.000 and a mean contact-frame count of 1.513. Occlusion sequences are balanced into 150 easy-occlusion and 150 hard-occlusion sequences.

Table A.2 summarizes the fixed-horizon scale/OOD evaluation split structure.

**Table A.2.** Dataset split counts for the fixed-horizon scale/OOD evaluation.

| Split | Count |
|---|---|
| train | 590 |
| val | 126 |
| test | 127 |
| test_id | 127 |
| test_ood_collision | 45 |
| test_ood_occlusion | 12 |

The OOD split is a controlled-generator parameter split, not a real-video transfer split. Collision OOD uses held-out high-angle/high-speed regimes; occlusion OOD uses longer hard-occlusion windows. Table A.3 gives the OOD definitions and summary statistics.

**Table A.3.** OOD split definitions and summary statistics.

| OOD item | Value |
|---|---|
| split_id | scale_ood_v0 |
| collision rule | top 15 percent by collision_angle_abs plus high approach speed |
| occlusion rule | hidden_duration >= 11 and hard occlusion |
| collision angle train mean | 2.4423 |
| collision angle OOD mean | 12.8934 |
| collision approach speed train mean | 2.4176 |
| collision approach speed OOD mean | 2.4625 |
| hidden duration train mean | 8.0644 |
| hidden duration OOD mean | 12.0000 |
| split overlap found | False |

## A.2 Model training

The fixed-horizon scale/OOD evaluation trains three passive object-state transition architecture families: GRU-R2h-loss-long, Transformer-lite full-pass configuration, and RSSM-large-position-heavy-long. The original evaluation used seeds 3407 and 3408; the final robustness extension adds independent seed 3409 under the same training protocol. Both kinematic_only and observation_like input variants are evaluated.

The kinematic_only variant uses normalized $x/y/v_x/v_y$ fields from the context window. The observation_like variant uses visible x/y, last observed x/y, and a visibility flag; hidden true object state is not provided as an input. Event labels, event phases, contact labels, pairwise relations, and future targets are not used as model context inputs. Table A.4 summarizes the shared training configuration.

**Table A.4.** Training configuration for the fixed-horizon transition models.

| Training item | Fixed-horizon value |
|---|---|
| architectures | GRU-R2h-loss-long; Transformer-lite full-pass configuration; RSSM-large-position-heavy-long |
| input variants | kinematic_only, observation_like |
| seeds | 3407, 3408, 3409 |
| optimizer | AdamW |
| learning rate | 0.001 |
| weight decay | 0.0001 |
| batch size | 96 |
| max epochs | 120 |
| early stopping patience | 24 |
| LR scheduler | ReduceLROnPlateau(mode='min', patience=12, factor=0.55) |
| gradient clipping | global norm clipped to 1.0 |
| loss weights | position=1.0, velocity=0.2, visibility=0.05, occlusion=0.05 |
| evaluation batch size | 128 |
| device | cuda if available else cpu |

The training objective is a weighted MSE over predicted future object state fields. The implemented loss is:

$$\text{loss} = 1.0 \times \text{position_MSE} + 0.2 \times \text{velocity_MSE} + 0.05 \times \text{visibility_MSE} + 0.05 \times \text{occlusion_MSE} \quad \text{(A.1)}$$

For every run, the training loop shuffles the training split with a seed-specific NumPy RNG, optimizes with AdamW, validates on the validation split each epoch, saves the best validation checkpoint, and reloads that checkpoint for downstream evaluation and diagnostics. The scheduler monitors validation position MSE. Training was configured for at most 120 epochs with early-stopping patience 24. In the completed fixed-horizon scale/OOD evaluation, all recorded training curves reached epoch 120: Table A.5 reports the completed runs and validation position MSE.

**Table A.5.** Completed training runs and validation position MSE.

| Variant | Architecture | Seed | Final epoch | Final val position MSE |
|---|---|---|---|---|
| kinematic_only | GRU-R2h-loss-long | 3407 | 120 | 0.000294534 |

| Variant | Architecture | Seed | Final epoch | Final val position MSE |
|---|---|---|---|---|
| kinematic_only | GRU-R2h-loss-long | 3408 | 120 | 0.000236122 |
| kinematic_only | RSSM-large-position-heavy-long | 3407 | 120 | 0.00012529 |
| kinematic_only | RSSM-large-position-heavy-long | 3408 | 120 | 0.000135432 |
| kinematic_only | Transformer-lite full-pass configuration | 3407 | 120 | 0.000117449 |
| kinematic_only | Transformer-lite full-pass configuration | 3408 | 120 | 0.000129377 |
| observation_like | GRU-R2h-loss-long | 3407 | 120 | 0.000297152 |
| observation_like | GRU-R2h-loss-long | 3408 | 120 | 0.000204312 |
| observation_like | RSSM-large-position-heavy-long | 3407 | 120 | 0.000162925 |
| observation_like | RSSM-large-position-heavy-long | 3408 | 120 | 0.000178298 |
| observation_like | Transformer-lite full-pass configuration | 3407 | 120 | 9.46372e-05 |
| observation_like | Transformer-lite full-pass configuration | 3408 | 120 | 0.000160981 |
| kinematic_only | GRU-R2h-loss-long | 3409 | 120 | 0.000258301 |
| kinematic_only | RSSM-large-position-heavy-long | 3409 | 120 | 0.000139615 |

| Variant | Architecture | Seed | Final epoch | Final val position MSE |
|---|---|---|---|---|
| kinematic_only | Transformer-lite full-pass configuration | 3409 | 120 | 9.53592e-05 |
| observation_like | GRU-R2h-loss-long | 3409 | 120 | 0.000257610 |
| observation_like | RSSM-large-position-heavy-long | 3409 | 120 | 0.000130009 |
| observation_like | Transformer-lite full-pass configuration | 3409 | 120 | 0.000156890 |

**A.3 Hidden-state extraction**

Hidden states are extracted during evaluation through the shared fixed-horizon evaluation function. For each batch, the model returns (prediction, embedding). The evaluation function accumulates predictions and embeddings without gradient tracking and computes position MSE against the future target positions.

The hidden-state definition is architecture-specific but standardized as the returned embedding:

- GRU-R2h-loss-long: the final recurrent hidden state of the top GRU layer.
- Transformer-lite full-pass configuration: the final-token encoder representation after the transformer encoder.
- RSSM-large-position-heavy-long: the latent state produced by the posterior/transition stack.

The extracted train embeddings are used for event-probe fitting. The extracted test/ID/OOD embeddings are used for event-probe evaluation, and diagnostic comparisons. Evaluation uses batch size 128 and runs on CUDA if available, otherwise CPU.

**A.4 Event probe**

The event probe is a lightweight linear classifier trained on model hidden states to classify event family: free motion, collision, or occlusion. It is used only as an event-regime readout diagnostic. It does not prove that the model causally uses the event representation.

The implemented probe standardizes train and test embeddings using the training embedding mean and standard deviation. Probe weights are initialized from a normal distribution with standard deviation 0.01, and the bias is initialized to zero. The classifier is optimized with full-batch gradient descent for 500 iterations with learning rate 0.08 and L2 weight penalty $10^{-4}$. Reported metrics include accuracy, macro-F1, balanced accuracy, majority baseline, and per-class precision/recall/F1.

The event probe is evaluated on the ID test split. Event-probe macro-F1 remains 1.000 for every kinematic-only architecture/seed row, including the added seed 3409. Table A.6 summarizes these sanity results.

**Table A.6.** Kinematic-only prediction, event-probe, and baseline diagnostic results.

| Variant | Architecture | Seed | ID_MSE | CV_MSE | Probe F1 | CFE collision |
|---|---|---|---|---|---|---|
| kinematic_only | GRU-R2h-loss-long | 3407 | 2.46E-04 | 3.35E-04 | 1 | 9.42E-05 |
| kinematic_only | Transformer-lite full-pass configuration | 3407 | 9.98E-05 | 3.35E-04 | 1 | 1.60E-04 |
| kinematic_only | RSSM-large-position-heavy-long | 3407 | 1.49E-04 | 3.35E-04 | 1 | 7.21E-04 |
| kinematic_only | GRU-R2h-loss-long | 3408 | 2.23E-04 | 3.35E-04 | 1 | 7.12E-05 |
| kinematic_only | Transformer-lite full-pass configuration | 3408 | 1.21E-04 | 3.35E-04 | 1 | 1.36E-04 |
| kinematic_only | RSSM-large-position-heavy-long | 3408 | 1.40E-04 | 3.35E-04 | 1 | 7.70E-04 |
| kinematic_only | GRU-R2h-loss-long | 3409 | 2.26E-04 | 3.35E-04 | 1 | 1.77E-03 |
| kinematic_only | Transformer-lite full-pass configuration | 3409 | 9.99E-05 | 3.35E-04 | 1 | 1.93E-03 |
| kinematic_only | RSSM-large-position-heavy-long | 3409 | 1.45E-04 | 3.35E-04 | 1 | 2.70E-03 |

### A.5 Original fixed-horizon CFE audit baseline and suppression sweep

The Causal Field Effect (CFE) measures causal sensitivity by modifying hidden representations and decoding predictions from the modified hidden state. Here and below, "field" in the CFE name and implementation identifiers refers only to an operational diagnostic/readout label; it does not assert that the model contains an internal physical field. In the original fixed-horizon baseline CFE score, event-specific examples are selected by mask, the model predicts the future target, and the baseline position MSE is measured. The hidden embedding is centered, partially suppressed, decoded through the model head, and evaluated again. In that original baseline implementation, CFE is defined as:

$$\text{CFE} = \text{perturbed_position_}MSE - \text{baseline_position_}MSE \quad \text{(A.2)}$$

Positive CFE means the perturbation increased event-relevant prediction error. This is causal sensitivity evidence, not complete mechanism identification.

The baseline implementation additionally performs a suppression-strength sweep. The strengths are 0.25, 0.5, 0.75, 1.0. The implemented sweep compares four intervention/control modes:

- target_field_suppression: suppresses the centered target embedding by the requested strength.
- matched_norm_random_direction: subtracts a random direction scaled to match the centered embedding standard deviation.
- field_label_shuffle: suppresses shuffled centered embeddings.
- random_subspace_control: suppresses a random binary subspace of the centered embedding.

The sweep is run for two diagnostic targets. CFE_collision uses the kinematic-only variant on the collision OOD split, and CFE_hard_occlusion uses the observation-like variant on the hard-occlusion OOD split. The original two-seed baseline sweep contains 192 rows. The seed-3409 extension adds 96 analogous rows under the same baseline sweep, giving 288 rows in the extended audit set. Table A.7 retains the original two-seed row allocation for audit history.

**Table A.7.** Suppression-sweep row allocation by diagnostic target and control.

| Target | Control | Rows |
|---|---|---|
| CFE_collision | field_label_shuffle | 24 |
| CFE_collision | matched_norm_random_direction | 24 |
| CFE_collision | random_subspace_control | 24 |
| CFE_collision | target_field_suppression | 24 |
| CFE_hard_occlusion | field_label_shuffle | 24 |
| CFE_hard_occlusion | matched_norm_random_direction | 24 |
| CFE_hard_occlusion | random_subspace_control | 24 |
| CFE_hard_occlusion | target_field_suppression | 24 |

The original baseline CFE sweep should be interpreted conservatively. It is useful as an audit baseline and as implementation history, but it is not the source definition for the Figure 2 results in the main text. Dose sensitivity in this baseline sweep supports causal diagnostic evidence, but controls such as field-label shuffle can be non-negligible because hidden-state factors are correlated. Therefore, the original baseline CFE sweep does not prove a complete causal mechanism.

### A.6 Relationship to the reported fixed-horizon projection CFE

The main CFE evidence reported in Section 6.3 and Figure 2 uses the reinforced fixed-horizon windowed projection CFE described mathematically in Appendix C. We evaluate explicit event windows within the prediction horizon, derive readout-aligned directions from a trained linear readout, and apply one-dimensional projection suppression in normalized hidden-state space. The final analysis uses three independent model seeds, with random-control seeds generated deterministically from SHA-256. All checkpoints, event windows, readout-aligned directions, suppression operators, alpha values, metrics, and interpretation rules are fixed across the analysis.

Code and data availability. The public release contains the training code, model definitions, diagnostic scripts, figure-generation scripts, dataset manifest, result tables, and non-primary audit materials used to support the manuscript. The detailed file-level mapping between manuscript results and repository artifacts is provided in the GitHub repository.

## Appendix B. Mathematical Details for Figure 1

This appendix defines the quantities plotted in Figure 1. Figure 1A, Figure 1D, and Figure 1E use hidden-state readout scores; Figure 1B and Figure 1C use data-level proxy z-scores; and Figure 1F uses positive shift fractions from a sliding-prefix analysis. The term field is retained here only as an operational readout label. These quantities should be compared within each Figure 1 subfigure, not across Figure 1 subfigures.

### B.1 Hidden-state readout scores for Figure 1A, Figure 1D, and Figure 1E

Let $h \in \mathbb{R}^d$ denote a hidden state from the model representation being analyzed. We use three structure labels: $K$ for kinematic, $C$ for contact, and $P$ for object-permanence. Let $\mathcal{F} = \{K, C, P\}$ denote this label set. Before applying the readout, the hidden state is standardized using the training-set mean and standard deviation used for probe training:

$$h_{\text{norm}} = \frac{h - \mu_{\text{train}}}{\sigma_{\text{train}}} \qquad \text{(B.1)}$$

The standardization is applied elementwise. The readout is a trained 3-class linear probe. It computes readout logits as

$$\ell = W h_{\text{norm}} + b, \qquad W \in \mathbb{R}^{3\times d}, \quad b \in \mathbb{R}^3 \qquad \text{(B.2)}$$

Here $W$ and $b$ are the learned weight matrix and bias vector of the linear probe, not parameters of the transition model. The three output dimensions of $\ell$ correspond to $K$, $C$, and $P$. Equivalently, for structure label $f$,

$$\ell_f = w_f^{\top} h_{\text{norm}} + b_f \qquad \text{(B.3)}$$

where $w_f$ is the row of $W$ associated with structure label $f$. The probe is trained on the training split with a cross-entropy objective and then frozen for the Figure 1 analyses.

The plotted hidden-state readout score is the softmax probability assigned to structure label $f$:

$$s_f(h) = \frac{\exp(\ell_f - m)}{\sum_{g\in\mathcal{F}} \exp(\ell_g - m)}, \qquad m = \max_{g\in\mathcal{F}} \ell_g \qquad \text{(B.4)}$$

The subtraction of $m$ is a numerically stable implementation of the standard softmax and does not change the probability. The scores over the three structure labels sum to one for each hidden state. Thus, $s_f(h)$ measures the relative probability mass assigned by the trained probe to structure label $f$ for that hidden state. It should be interpreted as a readout-assigned structure emphasis, not as an absolute physical intensity.

### Figure 1A aggregation

Figure 1A summarizes event-level readout emphasis. Let $\mathcal{I}_e$ be the set of held-out sequences belonging to event family $e$, and let $\mathcal{T}_i$ denote the set of hidden states from sequence $i$ included in the event-level aggregation. The event-level score for structure label $f$ is

$$S_{e,f} = \frac{1}{|\mathcal{I}_e|} \sum_{i\in\mathcal{I}_e} \left( \frac{1}{|\mathcal{T}_i|} \sum_{t\in\mathcal{T}_i} s_f\left(h_{i,t}\right) \right) \qquad \text{(B.5)}$$

If a sequence-level or final hidden representation is used, $\mathcal{T}_i$ contains that representation. If multiple time entries are used, the probability is first averaged within sequence/time entries and then averaged across sequences in the same event family.

**Figure 1D and E aligned-time aggregation**

Figure 1D and Figure 1E summarize time-aligned hidden-state dynamics. For a sequence $i$, let $b_i$ denote the event boundary used for alignment: the contact onset for collision and hidden-start for occlusion. For relative time $\tau$, the aligned readout score is

$$S_f(\tau) = \frac{1}{|\mathcal{I}(\tau)|} \sum_{i \in \mathcal{I}(\tau)} s_f\left(h_{i,b_i+\tau}\right) \tag{B.6}$$

where $\mathcal{I}(\tau)$ is the set of sequences for which the aligned time $b_i + \tau$ is available. When multiple seeds are plotted, the curve is computed per seed and then summarized by the mean across seeds; error bars, when shown, represent the corresponding standard error of the mean.

**B.2 Data-level proxy z-scores for Figure 1 B and Figure 1C**

Figure 1 B and Figure 1C are not hidden-state readouts. They describe the physical demands of annotated event phases using data-level proxies derived from object states, pairwise relations, event-phase labels, and visibility/contact annotations. Kinematic proxies summarize motion-related quantities such as displacement, velocity, or visible trajectory structure. Contact proxies summarize pairwise interaction or contact-related structure. Object-permanence proxies summarize occlusion, hidden-state, or reappearance-related demands.

Let $\rho_{f,\phi}$ denote the raw data-level proxy value for structure label $f$ in phase $\phi$. For Figure 1B, phases include pre-contact, contact, and post-contact. For Figure 1C, phases include pre-occlusion, hidden, and reappearance when that phase is available in the selected analysis table. For each of Figure 1B and Figure 1C, the raw structure-by-phase proxy matrix is standardized separately. Let $\Phi$ denote the phase set used in the corresponding figure:

$$\mu_{\text{proxy}} = \frac{1}{|\mathcal{F}||\Phi|} \sum_{f \in \mathcal{F}} \sum_{\phi \in \Phi} \rho_{f,\phi}, \tag{B.7}$$

$$\sigma_{\text{proxy}} = \sqrt{\frac{1}{|\mathcal{F}||\Phi|} \sum_{f \in \mathcal{F}} \sum_{\phi \in \Phi} \left(\rho_{f,\phi} - \mu_{\text{proxy}}\right)^2}, \tag{B.8}$$

$$\zeta_{f,\phi} = \frac{\rho_{f,\phi} - \mu_{\text{proxy}}}{\sigma_{\text{proxy}}} \tag{B.9}$$

Thus, $\zeta_{f,\phi}$ indicates whether a proxy-defined structure demand is relatively high or low within the corresponding Figure 1B or Figure 1C analysis. These proxy z-scores are used to interpret the physical structure of the trajectory itself. They do not imply that the model hidden state activates the corresponding structure. Here $\mu_{proxy}$ and $\sigma_{proxy}$ are computed separately for Figure 1B and Figure 1C; $\zeta_{f,\phi}$ denotes the resulting standardized proxy score.

**B.3 Positive shift fractions and directional consistency for Figure 1F**

Figure 1F summarizes the lighter sliding-prefix hidden-state analysis for GRU and Transformer-lite. Unlike the RSSM rollout-latent analysis in Figure 1D and Figure 1E, this analysis does not reconstruct full phase-resolved hidden dynamics. It only asks whether the expected readout score changes in the expected direction for each evaluated case.

Let $\mathcal{R}$ denote the set of evaluated cases, such as architecture–seed–variant combinations. For each $r \in \mathcal{R}$, hidden states are extracted from sliding prefixes, and the same readout-score logic is applied to the relevant prefix-conditioned hidden states. Here, $r$ indexes individual evaluated cases.

For collision, the expected effect is a positive shift in the contact score around the contact/collision phase relative to an earlier baseline phase. Define

$$\bar{s}^r_{C,\text{target}} = \text{mean}_{(i,t)\in\Omega^r_{\text{contact}}} s_C\left(h^r_{i,t}\right) \tag{B.10}$$

$$\bar{s}^r_{C,\text{base}} = \text{mean}_{(i,t)\in\Omega^r_{\text{pre-contact}}} s_C\left(h^r_{i,t}\right) \tag{B.11}$$

and

$$\Delta^r_C = \bar{s}^r_{C,\text{target}} - \bar{s}^r_{C,\text{base}} \tag{B.12}$$

For occlusion, the expected effect is a positive shift in the object-permanence score during the hidden/occlusion phase relative to a pre-occlusion baseline. Define

$$\bar{s}^r_{P,\text{target}} = \text{mean}_{(i,t)\in\Omega^r_{\text{hidden}}} s_P\left(h^r_{i,t}\right) \tag{B.13}$$

$$\bar{s}^r_{P,\text{base}} = \text{mean}_{(i,t)\in\Omega^r_{\text{pre-occlusion}}} s_P\left(h^r_{i,t}\right) \tag{B.14}$$

and

$$\Delta^r_P = \bar{s}^r_{P,\text{target}} - \bar{s}^r_{P,\text{base}} \tag{B.15}$$

Here $C$ denotes the contact structure label and $P$ denotes the object-permanence structure label. The windows $\Omega^r_{\text{contact}}$, $\Omega^r_{\text{hidden}}$, $\Omega^r_{\text{pre-contact}}$, and $\Omega^r_{\text{pre-occlusion}}$ follow the phase labels available in the sliding-prefix analysis tables.

For each evaluated case, we convert the two shifts into directional indicators:

$$I^r_C = \mathbf{1}[\Delta^r_C > 0], \qquad I^r_P = \mathbf{1}[\Delta^r_P > 0] \tag{B.16}$$

For an architecture group $A \subseteq R$, the positive shift fractions are

$$\text{PSF}^C_A = \frac{1}{|A|}\sum_{r\in A} I^r_C, \qquad \text{PSF}^P_A = \frac{1}{|A|}\sum_{r\in A} I^r_P \tag{B.17}$$

The joint directional consistency score aggregates the two expected shifts across all evaluated cases:

$$\mathrm{DCS} = \frac{1}{2|\mathcal{R}|} \sum_{r \in \mathcal{R}} (I_C^r + I_P^r) \tag{B.18}$$

Figure 1F reports these positive shift fractions and the joint DCS. A value of 1.000 means that every evaluated case showed the expected positive direction. This supports directional consistency in prefix-conditioned hidden states, but it should not be interpreted as full time-resolved rollout dynamics or as evidence of explicit physical modules.

### B.4 Interpretation boundary

The Figure 1 quantities have different scales and should not be compared across Figure 1A-F. Figure 1A, Figure 1D, and Figure 1E report hidden-state readout probabilities; Figure 1B and Figure 1C report standardized data-level proxy values; and Figure 1F reports positive-shift fractions. The intended interpretation is event-conditioned organization of overlapping diagnostic structure, not one-to-one event modules or a complete mechanistic decomposition.

## Appendix C. Mathematical and Implementation Details for Causal Field Effects

This appendix defines the causal field effect (CFE) quantities used in Section 6.3. Here, "field" is retained only in the established CFE name and implementation labels; the scientific object being tested is readout-aligned latent structure. The reported implementation uses fixed-horizon, windowed projection CFE. Event windows are evaluated only where they overlap the configured future prediction horizon. This means that CFE is computed only on frames for which the base model and the perturbed model both produce predictions.

### C.1 Notation and fixed-horizon event windows

Let $r$ index an evaluated case, such as an architecture-seed-variant combination. Let $i$ index a sequence, $t$ index a predicted frame, $o$ index an object, and $j \in \{x, y\}$ index the position coordinate used in the CFE loss. The transition model produces a hidden representation $h_i^r$ or $h_{i,t}^r$, depending on the architecture-specific interface, and a prediction $\hat{y}_{i,t,o,j}^r$ for the target state $y_{i,t,o,j}$. Here r is the evaluated-case index and j is the x/y coordinate index; d remains reserved for hidden-state dimension.

For compact notation, let

$$\mathcal{F} = \{K, C, P\} \tag{C.1}$$

where $K$ denotes the kinematic structure label, $C$ denotes the contact structure label, and $P$ denotes the object-permanence structure label.

Let $T_{\text{future}}$ denote the fixed set of frames predicted by the model. In the reported experiments, the model observes the first eight frames and predicts the next eight frames, so the prediction horizon corresponds to the configured future target frames. Let $\tilde{\Omega}_e(i)$ denote the event-defined raw window for event family $e$ in sequence $i$. The implemented CFE window is the intersection of the event-defined window and the predicted future horizon: We use Ω for event windows so that W is reserved for the linear-readout weight matrix.

$$\Omega_e(i) = \tilde{\Omega}_e(i) \cap T_{\text{future}} \tag{C.2}$$

Equivalently, the implemented window contains exactly those frames that are both event-relevant and predicted by the model:

$$t \in \Omega_e(i) \quad \Leftrightarrow \quad t \in \tilde{\Omega}_e(i) \text{ and } t \in T_{\text{future}} \tag{C.3}$$

For collision, $\Omega_e(i)$ is centered on the contact regime after this intersection. For hard occlusion, the primary CFE window covers hidden frames within the predicted horizon. The current primary analysis does not report a reappearance-window CFE, because no reappearance frames remain in the configured fixed-horizon CFE window.

### C.2 Base event-window prediction loss

The base prediction loss is computed without intervention. The reported CFE loss is normalized $x/y$ position MSE over the selected sequence-frame-object-coordinate entries. For an event family $e$ and case

$r$, define the set of evaluated sequences as $I_e^r$. Let

$$N_e^r = \sum_{i \in I_e^r} \sum_{t \in \Omega_e(i)} \sum_o \sum_{j \in \{x,y\}} 1 \quad \text{(C.4)}$$

be the number of evaluated sequence-frame-object-coordinate entries. The base event-window loss is

$$L_{\text{base}}^r(e) = \frac{1}{N_e^r} \sum_{i \in I_e^r} \sum_{t \in \Omega_e(i)} \sum_o \sum_{j \in \{x,y\}} \left(\hat{y}_{i,t,o,j}^r - y_{i,t,o,j}\right)^2 \quad \text{(C.5)}$$

Thus, the base loss is not a full-state loss over all target channels. It is the normalized position-prediction loss in the fixed-horizon event window.

**C.3 Readout-aligned direction construction from the linear readout**

The readout-aligned directions used for projection suppression are derived from the trained 3-class linear readout. Given a hidden representation $h$, we first apply the same training-set normalization used by the readout:

$$z = \frac{h - \mu_{\text{train}}}{\sigma_{\text{train}}} \quad \text{(C.6)}$$

The readout produces logits

$$\ell = Wz + b \quad \text{(C.7)}$$

where $W \in \mathbb{R}^{3 \times d}$ is the learned weight matrix, $b \in \mathbb{R}^3$ is the learned bias vector, and the three output dimensions correspond to $K$, $C$, and $P$. Let $w_f$ denote the row of $W$ associated with structure label $f$. For the reported probe_logit_contrast direction, we compute

$$v_f = w_f - \frac{1}{|\mathcal{F}|-1} \sum_{g \in \mathcal{F},\, g \neq f} w_g \quad \text{(C.8)}$$

The unit readout-aligned direction is

$$u_f = \frac{v_f}{\|v_f\|_2} \quad \text{(C.9)}$$

This direction should be interpreted as a readout-aligned representational direction, not as a unique physical module. In a multi-class readout, readout evidence is relative, and readout directions can overlap with other predictive structure.

**C.4 Projection suppression intervention**

Let $A_{f,\alpha}$ denote the intervention that suppresses the component of the normalized hidden representation associated with structure label $f$ at suppression strength $\alpha$. The reported sweep evaluates

$$\alpha \in \{0.25, 0.50, 0.75, 1.00\} \quad \text{(C.10)}$$

For one-dimensional projection suppression, the perturbed normalized representation is

$$z^{(-f,\alpha)} = A_{f,\alpha}(z) = z - \alpha \left(z^\top u_f\right) u_f \quad \text{(C.11)}$$

The operator $A_{f,\alpha}$ should be interpreted as a representational-subspace intervention. It suppresses a readout-aligned direction; it does not prove that the model contains an explicit physical module or isolated causal circuit.

The perturbed representation is then mapped back to the model's hidden-state scale using the inverse of the readout normalization:

$$h^{(-f,\alpha)} = \mu_{\text{train}} + \sigma_{\text{train}} \odot z^{(-f,\alpha)} \quad \text{(C.12)}$$

### C.5 Perturbed prediction loss and CFE

After intervention, the perturbed hidden representation is decoded to obtain perturbed predictions $\hat{y}_{i,t,o,j}^{r,(-f,\alpha)}$. For GRU and Transformer-lite, the perturbed representation is decoded through the corresponding prediction head. For RSSM-lite in the reported analysis, CFE is computed by head decoding from the perturbed latent state; it is not a full RSSM transition re-rollout intervention.

The perturbed event-window loss is

$$L_{\text{pert}}^{r}(e,f,\alpha) = \frac{1}{N_e^r} \sum_{i \in I_e^r} \sum_{t \in \Omega_e(i)} \sum_o \sum_{j \in \{x,y\}} \left( \hat{y}_{i,t,o,j}^{r,(-f,\alpha)} - y_{i,t,o,j} \right)^2 \quad \text{(C.13)}$$

The causal field effect is the event-window loss increase caused by the readout-aligned intervention:

$$\text{CFE}^r(e,f,\alpha) = L_{\text{pert}}^r(e,f,\alpha) - L_{\text{base}}^r(e) \quad \text{(C.14)}$$

A positive CFE means that suppressing the readout-aligned component worsens prediction in the evaluated event window. Let $e_{\text{col}}$ denote the collision-contact window and $e_{\text{hid}}$ denote the hard-occlusion hidden window. The two primary CFE quantities used in Section 6.3 are

$$\text{CFE}_{\text{col}}^r(\alpha) = \text{CFE}^r(e_{\text{col}}, C, \alpha) \quad \text{(C.15)}$$

$$\text{CFE}_{\text{hid}}^r(\alpha) = \text{CFE}^r(e_{\text{hid}}, P, \alpha) \quad \text{(C.16)}$$

The current primary analysis should not be described as reappearance-window CFE. When CFE is reported without specifying a suppression-strength sweep, it refers to the configured suppression strength for that summary, typically $\alpha = 1$. Aggregate target effects below use the case-averaged CFE defined in Eq. (C.17).

### C.6 Aggregation across cases

For a set of evaluated cases $\mathcal{R}$, such as all architecture-seed combinations, the mean CFE at event $e$, target structure label $f$, and suppression strength $\alpha$ is

$$\overline{\text{CFE}}(e,f,\alpha) = \frac{1}{|\mathcal{R}|} \sum_{r \in \mathcal{R}} \text{CFE}^r\,(e,f,\alpha) \quad \text{(C.17)}$$

When plotted, error bars or standard errors can be computed over the same set of cases. In the main text, the sign of the effect, target-control comparison, and dose trend are more important than the absolute magnitude of any single loss delta.

### C.7 Matched controls and target-control margins

CFE is interpreted relative to matched controls. Let q denote a control intervention type. The perturbed control loss uses the same event-window loss as Eq. (C.13), with control intervention q replacing the target readout-aligned suppression. The case-level control CFE is

$$\mathrm{CFE}_q^r(e,f,\alpha) = L_{\mathrm{pert},q}^r(e,f,\alpha) - L_{\mathrm{base}}^r(e) \tag{C.18}$$

The main controls are matched-norm random direction, random subspace control, and field-label shuffle. For the final three-seed analysis, random-control RNG seeds are derived deterministically as $seed_{ctrl} = seed_{model} + offset_{SHA256}(vversion, architecture, event, direction\ method)$, where the offset is deterministically computed from a versioned SHA-256 hash of the architecture, event, and direction method.

A useful target-control margin is

$$M_q(e,f,\alpha) = \frac{1}{|\mathcal{R}|}\sum_{r\in\mathcal{R}}\left[\mathrm{CFE}^r(e,f,\alpha) - \mathrm{CFE}_q^r(e,f,\alpha)\right] \tag{C.19}$$

Equation (C.19) averages the paired target-minus-control CFE over the same evaluated cases. This is equivalent to subtracting the case-averaged control CFE from the case-averaged target CFE. Positive margins indicate that target suppression degrades prediction more than the corresponding control. The interpretation is comparative: a positive target CFE without positive margins may indicate general perturbation sensitivity rather than target-specific functional relevance. Conversely, negative or weak margins should be interpreted as reduced specificity, not as evidence for an explicit module.

### C.8 Suppression-strength sweep

The suppression sweep evaluates whether CFE increases as the intervention becomes stronger. For strengths $\alpha_1 < \alpha_2$, a dose-sensitive pattern satisfies

$$\mathrm{CFE}^r(e,f,\alpha_2) \geq \mathrm{CFE}^r(e,f,\alpha_1), \qquad r\in\mathcal{R} \tag{C.20}$$

for most adjacent strength pairs and evaluated cases. Perfect monotonicity is not required, because finite samples, different architectures, and overlapping readout-aligned directions can introduce deviations. The relevant pattern is a broadly dose-sensitive increase in event-window prediction loss under stronger target suppression, interpreted together with matched controls.

### C.9 Interpretation boundary

CFE is a functional relevance measure within the configured prediction setting, not a full causal-mechanistic explanation. It supports the claim that readout-aligned representational structure has predictive consequences in the relevant fixed-horizon event window. It does not establish that the model contains an explicit physical module, that the intervention isolates a unique circuit, that the RSSM transition has been fully re-rolled under intervention, or that the result generalizes to arbitrary sliding-context forecasts.

The strongest supported interpretation is therefore: readout-aligned components are functionally relevant when suppressing them selectively increases prediction loss in the fixed-horizon event window where that structure is expected to matter, relative to matched controls.

### C.10 Summary of the empirical pattern

The core expected pattern is positive collision-contact CFE and positive hard-occlusion hidden-window CFE, together with appropriate controls:

$$\overline{\mathrm{CFE}}(e_{\mathrm{col}}, C, \alpha) > 0 \tag{C.21}$$

$$\overline{\mathrm{CFE}}(e_{\mathrm{hid}}, P, \alpha) > 0 \tag{C.22}$$

Reappearance-window CFE is not part of the current primary result. In the three-seed stable-RNG analysis, collision-contact shows positive target-control margins across all reported controls. Hard occlusion shows positive sensitivity but mixed specificity across controls. Together with broadly dose-sensitive target effects, these results support fixed-horizon functional sensitivity of contact- and object-permanence-aligned structure without implying explicit modules, isolated causal circuits, full RSSM rollout interventions, or arbitrary sliding-context generalization.